\documentclass[10pt,twocolumn,letterpaper]{article}

\usepackage{wacv}              

\usepackage{graphicx}
\usepackage{amsmath}
\usepackage{amssymb}
\usepackage{booktabs}
\usepackage{multirow}
\usepackage{soul}
\usepackage[accsupp]{axessibility}
\usepackage[normalem]{ulem}
\useunder{\uline}{\ul}{}

\usepackage[pagebackref,breaklinks,colorlinks]{hyperref}

\usepackage[capitalize]{cleveref}
\crefname{section}{Sec.}{Secs.}
\Crefname{section}{Section}{Sections}
\Crefname{table}{Table}{Tables}
\crefname{table}{Tab.}{Tabs.}

\def\wacvPaperID{13} 
\def\confName{WACV}
\def\confYear{2025}

\begin{document}

\title{MADation: Face Morphing Attack Detection with Foundation Models}

\author{Eduarda Caldeira$^{1}$, Guray Ozgur$^{1}$, Tahar Chettaoui$^{1}$, Marija Ivanovska$^{2}$,\\Peter Peer$^{2}$, Fadi Boutros$^{1}$, Vitomir Struc$^{2}$, Naser Damer$^{1,3}$ \vspace{2mm} \\ 
$^{1}$ Fraunhofer IGD, Germany, $^{2}$ University of Ljubljana, Slovenia\\
$^{3}$ TU Darmstadt, Germany \\ 
Email: maria.eduarda.loureiro.caldeira@igd.fraunhofer.de
}

\maketitle
\begin{abstract}
    Despite the considerable performance improvements of face recognition algorithms in recent years, the same scientific advances responsible for this progress can also be used to create efficient ways to attack them, posing a threat to their secure deployment. Morphing attack detection (MAD) systems aim to detect a specific type of threat, morphing attacks, at an early stage, preventing them from being considered for verification in critical processes. Foundation models (FM) learn from extensive amounts of unlabelled data, achieving remarkable zero-shot generalization to unseen domains. Although this generalization capacity might be weak when dealing with domain-specific downstream tasks such as MAD, FMs can easily adapt to these settings while retaining the built-in knowledge acquired during pre-training. In this work, we recognize the potential of FMs to perform well in the MAD task when properly adapted to its specificities. To this end, we adapt FM CLIP architectures with LoRA weights while simultaneously training a classification header. The proposed framework, MADation surpasses our alternative FM and transformer-based frameworks and constitutes the first adaption of FMs to the MAD task. MADation presents competitive results with current MAD solutions in the literature and even surpasses them in several evaluation scenarios. To encourage reproducibility and facilitate further research in MAD, we publicly release the implementation of MADation at https://github.com/gurayozgur/MADation
\end{abstract}

\vspace{-6mm}
\section{Introduction}
\vspace{-2mm}
The high focus of the research community on the study of deep learning techniques in recent years has led to the development of high-performing systems in several fields, including face recognition (FR) \cite{boutros2022elasticface, deng2019arcface}. However, the same scientific advances used to improve the recognition power of FR systems can also be used to create efficient ways to attack them \cite{damer2018morgan, DBLP:conf/icb/FerraraFM14}, posing a threat to their secure deployment. Morphing attacks (MA) constitute an example of such threats, as their generation process aims at incorporating features from more than one identity, resulting in a sample that can be verified by multiple people by the same FR system. When left undetected, these attacks can lead to several dangerous situations \cite{damer2023mordiff, caldeira2023unveiling, DBLP:journals/tbbis/ZhangVRRDB21}, such as allowing multiple people to pass border control with the same passport or letting 
a criminal 
travel under the identity of another person \cite{DBLP:conf/visapp/MakrushinND17}. To address this problem, several morphing attack detection (MAD) systems have been proposed \cite{huber2022syn, caldeira2023unveiling, fang2022unsupervised, damer2021pw, neto2022orthomad, ramachandra2019detecting}. MAD algorithms aim at distinguishing unaltered images (bona-fide samples) from MAs to identify malicious samples at an early stage and prevent them from being considered for verification in critical processes. 

Foundation models (FM) are large-scale networks that can be trained with self-supervised learning, which allows them to learn from unlabelled data. The fact that no labelling is required for FMs' training samples highly simplifies the data acquisition task, allowing FMs to be trained in massive and diverse datasets. This training paradigm results in models that can efficiently generalize to a wide variety of assignments \cite{DBLP:journals/corr/abs-2108-07258}, making them particularly beneficial for fields that address several tasks, such as natural language processing (NLP) \cite{brown2020language} and computer vision (CV) \cite{kirillov2023segment, ravi2024sam, oquab2023dinov2, radford2021learning}. Despite the recent attention given to FMs, their adaption to perform biometrics tasks is still very limited. While very recent works have used FMs to generate synthetic face images \cite{papantoniou2024arc2face}, perform iris segmentation \cite{farmanifard2024iris} and FR \cite{chettaoui2024froundation}, the utility of FMs for most biometrics fields is still highly under-explored. This literature gap should be carefully addressed, especially taking into account that biometrics tasks such as MAD may strongly benefit from FMs' high generalization power, provided their efficient adaption to the domain specificities of the downstream task \cite{chettaoui2024froundation}, and high generalizability when it comes to sub-domains \cite{radford2021learning}, i.e. different morphing mechanisms in the MAD case.

This work explores the potential of using FMs as the basis for the downstream MAD task. Given the domain-specific nature of the MAD task and knowing that FMs often underperform in specialized settings \cite{DBLP:journals/corr/abs-2308-07156}, we propose to adapt the pre-trained FM CLIP \cite{radford2021learning} to MAD with low-rank adaption (LoRA), while simultaneously training a header to perform classification. This allows the FM to better align its feature space with the specificities of the downstream MAD task and still take advantage of the built-in knowledge acquired during pre-training. This adaption paradigm corresponds to our proposed framework, MADation. We further evaluate whether MADation properly takes advantage of FMs' properties by comparing it with alternative FM and transformer-based methods. We start by assessing the importance of adapting CLIP to the downstream MAD task by evaluating its zero-shot performance on this task (TI). Then, to verify whether LoRA adaption improves MAD performance, we use the FM as a frozen feature extractor. In this scenario, the FM's feature space is not aligned with the specificities of MAD and only the classification layer is trained (FE). Finally, to ensure that MADation's performance is not only deriving from its architecture but is also dependent on the FM's built-in knowledge acquired during pre-training, we assess the importance of the FM's pre-trained weights by comparing MADation with models following the same architectures trained from scratch (ViT-FS). The developed experiments highlight the efficiency of MADation compared with the remaining methods, reducing the average EER by 16.93 pp. and 8.10 pp. compared to ViT-FS and FE, respectively, for CLIP ViT-L. Furthermore, MADation revealed competitive performance levels compared with recent MAD solutions, highlighting FM's potential in domain-specific tasks such as MAD.

\vspace{-2mm}
\section{Related Work} \label{sec:sota}
\begin{figure*}[tbh!]
\centering
\includegraphics[width=0.77\textwidth,keepaspectratio]{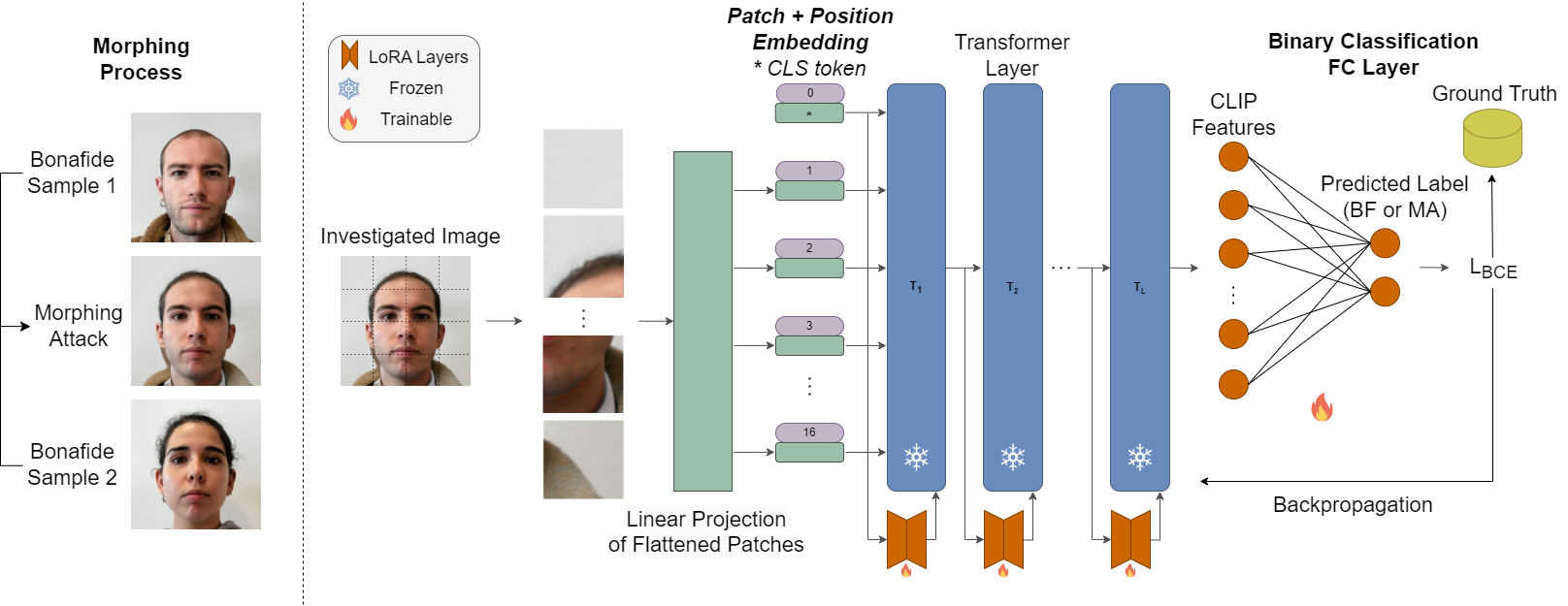}
\vspace{-5mm}
    \caption{Morphing attack generation and MADation's pipeline. The left side of the figure depicts a morphing sample and the two bona-fide identities that were morphed to generate it, using \cite{damer2023mordiff}. Keep in mind that attackers commonly choose to morph faces with similar features for higher success \cite{DBLP:conf/icb/DamerSZWTKK19}. The right side represents MADation's pipeline, consisting of an adapted FM followed by a binary fully connected classification layer. The embedding space of the FM is adapted by fine-tuning the LoRA parameters and the classification layer is simultaneously trained to produce the MAD predictions. Better visualized in colour.}
    \label{fig:pipeline}
    \vspace{-6mm}
\end{figure*}

\vspace{-2mm}
\textbf{MAD:} MAs are face images that result from the fusion of identity information belonging to two or more identities, allowing them to be simultaneously verifiable as belonging to all of them. An example of an MA can be found in Figure \ref{fig:pipeline}. Both human observers and FR systems are vulnerable to MAs \cite{ferrara2016effects, scherhag2017vulnerability}, leading to dangerous situations, such as multiple people being able to pass border control with the same passport \cite{caldeira2023unveiling}. To address the threat posed by MAs, several studies have proposed MAD systems \cite{huber2022syn, caldeira2023unveiling, fang2022unsupervised, damer2021pw, neto2022orthomad, ramachandra2019detecting}. These systems can address the MAD task from two different perspectives, depending on the MAD operational scenario. Differential MAD solutions \cite{damer2019detecting} are fed two samples simultaneously: a live capture of the individual claiming that the investigated image represents their identity and the investigated image itself. Although this approach is useful in scenarios such as border control, the fact that two images need to be compared to perform the detection limits its applicability in several scenarios \cite{damer2019detecting}, e.g. analysing stand-alone documents. Hence, several studies have developed single-image MAD systems \cite{huber2022syn, caldeira2023unveiling, fang2022unsupervised, damer2021pw, neto2022orthomad, ramachandra2019detecting, DBLP:conf/iwbf/IvanovskaS23, zhang2024generalized}, which can detect whether the investigated image is a morph based only on its characteristics. Ramachandra~\textit{et al.}~\cite{ramachandra2019detecting} introduced a handcrafted-feature-based approach that extracts textural features across multiple scales and classifies them using collaborative representation. \cite{damer2021pw} deviated from the common binary classification of the whole investigated image by learning to classify each of its pixels (or pixel blocks) as bona-fide or MA. Fang~\textit{et al.}~\cite{fang2022unsupervised} proposed an unsupervised approach that used self-paced learning to assign smaller weights to suspicious samples, which generally correspond to MAs, allowing for training a robust autoencoder for anomaly detection even when the training data is polluted with MAs. Neto~\textit{et al.}~\cite{neto2022orthomad} determined whether the analyzed sample contained two independent identities by separating its identity information into two orthogonal latent vectors. \cite{caldeira2023unveiling} trained an autoencoder on bona-fide samples to distil identity knowledge to a MAD system, following distinct distillation techniques for bona-fide and MAs. \cite{zhang2024generalized} performed MAD with ViT architectures, showing promising results. \cite{DBLP:conf/iwbf/IvanovskaS23} developed a self-supervised diffusion model that reconstructs bona-fide images from noisy inputs. As the model is trained on bona-fide samples alone, it leads to higher error rates when fed MAs, which can be identified through anomaly detection. \cite{huber2022syn} promoted the SYN-MAD 2022 competition on MAD based on synthetic training data, presenting a comprehensive analysis of the results of seven submitted approaches. This work also focuses on single-image MAD due to its wider utility in real-world scenarios, whether they offer a live probe or not. 

\vspace{-1mm}
\textbf{Foundation Models:} FMs contain many trainable parameters, enabling FM to learn from large and diverse datasets.
This intensive training leads to high adaptability, which is particularly useful in areas that deal with a wide range of tasks, such as CV \cite{kirillov2023segment, radford2021learning, farmanifard2024iris, oquab2023dinov2, papantoniou2024arc2face, chettaoui2024froundation}. DINOv2 networks \cite{oquab2023dinov2} is a series of self-supervised pre-trained visual models able to generate universal features that can be used to perform both image-level and pixel-level visual tasks. The Segment Anything Model (SAM) \cite{kirillov2023segment} can perform image segmentation on a wide range of domains. 
Contrastive Language-Image Pretraining (CLIP)  \cite{radford2021learning} is a multimodal FM trained to interpret visual and textual inputs simultaneously, effectively learning the correlation between images and their textual description. 

Although vision FMs' prove to be generalizable for several downstream tasks,  they achieve less optimal performance when applied to some specific settings \cite{DBLP:journals/corr/abs-2308-07156}, which require adapting FMs to the desired downstream task \cite{chen2022adaptformer, chen2022vision, hu2021lora}. AdaptFormer \cite{chen2022adaptformer} used two identical MLP branches instead of the MLP block in the transformer encoder. One of them replicates the original network to help maintain its properties, while the other allows for task-specific fine-tuning. ViT-Adapter \cite{chen2022vision} led to state-of-the-art (SOTA) COCO results for plain ViT networks, by combining the reconstruction of fine-grained multi-scale features with the introduction of image-related inductive biases into the FM. Hu~\textit{et al.}~\cite{hu2021lora} inserted trainable rank decomposition matrices in the pre-trained FM, allowing for low-dimensional reparametrization. During the adaption process, the FM's pre-trained weights are kept frozen and only these matrices' weights are updated, allowing for an effective adaption with minimal computational cost. Using these matrices, also known as Low-Rank Adaptation (LoRA) layers, to adapt FMs has led to performance improvements in a broad range of tasks, such as capsule endoscopy diagnosis \cite{zhang2024learning}, plant phenotyping \cite{chen2023adapting} and FR \cite{chettaoui2024froundation}. In this work, we select LoRA to adapt CLIP to the MAD task due to its promising results reported in the literature, namely in biometrics \cite{chettaoui2024froundation}.

Despite the growing attention given to FMs by the community, their application in biometrics is still limited to a few recent works \cite{farmanifard2024iris, papantoniou2024arc2face, chettaoui2024froundation}. In \cite{farmanifard2024iris}, SAM \cite{kirillov2023segment} was fine-tuned to iris segmentation. \cite{papantoniou2024arc2face} used FMs to generate identity-specific facial images. A recent work \cite{chettaoui2024froundation} used LoRA \cite{hu2021lora} to fine-tune DINOv2 \cite{oquab2023dinov2} and CLIP \cite{radford2021learning} FMs to the FR task under several data availability settings. The experiments showed that FMs can be efficiently used in FR, especially in low data availability scenarios, where the proposed technique surpassed models trained from scratch. 

In this work, we contribute to the list of recent advancements in FMs' application to biometrics by proposing a MAD solution based on FMs, MADation. In particular, we recognize the FMs' generalization potential and their utility to tasks such as MAD when properly adapted to their domain-specific characteristics \cite{chettaoui2024froundation}. To this end, MADation incorporates LoRA layers in the analysed FM while simultaneously training an extra classification layer.
\vspace{-2mm}
\section{Methodology} \label{sec:methodology}
\vspace{-1mm}
\subsection{Preliminary on CLIP} \label{sec:clip}
\vspace{-2mm}

CLIP \cite{radford2021learning} is a multimodal FM that interprets visual and textual inputs simultaneously. CLIP was trained in a large dataset where each image is paired with a textual description, allowing it to learn the relationship between these two modalities. When a pair is fed to CLIP, its components are processed by two distinct encoders that are simultaneously trained using a contrastive learning approach that evaluates the cosine similarity between the features extracted for image and text, maximizing or minimizing it for positive and negative pairs, respectively. This allows CLIP to effectively learn the correlation between images and textual descriptions, resulting in a model generalizable across distinct tasks \cite{radford2021learning} with very competitive zero-shot learning results. 

In this work, we use CLIP due to its capacity to generalize well to domain-specific downstream tasks when properly adapted \cite{chettaoui2024froundation}. We assess CLIP's abilities as a MAD model by evaluating its zero-shot learning performance (Section \ref{sec:baseline}). To this end, CLIP is fed with image-text pairs where the text input corresponds to the possible image labels (`face image morphing attack' and `bona-fide presentation' based on the ISO/IEC 20059 standard \cite{ISO20059}). For the remaining approaches, only the image encoder is used and CLIP works as a feature extractor on top of which a classification layer is added, following the recent successful utilization of FMs in downstream tasks such as FR \cite{chettaoui2024froundation} and image segmentation \cite{kirillov2023segment}. These approaches include our proposed framework, MADation (Section \ref{sec:MADation}), and two additional frameworks designed to assess the importance of the FM's built-in knowledge and of adapting its feature space to the downstream domain (Section \ref{sec:baseline}). When applicable (TI, FE, and MADation), CLIP was initialized with the pre-trained weights made publicly available in \cite{radford2021learning}\footnote{\url{https://github.com/OpenAI/CLIP}}.

\vspace{-1mm}
\subsection{MADation} \label{sec:MADation}
\vspace{-2mm}
In this work, we take advantage of the high generalization capacity of FMs and assess their usefulness in the downstream MAD task. Although we acknowledge the benefits that can arise from directly deploying FMs to MAD, we also recognize the limited ability of these networks to perform well in domain-specific tasks such as MAD without any adaption. Thus, the proposed framework, MADation, adapts an FM by using LoRA layers to shift its pre-trained feature space in a direction that facilitates the MAD task while simultaneously training a classification layer. A visual representation of MADation is depicted in Figure \ref{fig:pipeline}.

\textbf{Fine-Tuning with LoRA:} As mentioned in Section \ref{sec:sota}, the FMs' drop in performance when facing domain-specific scenarios such as MAD can be addressed with ViT adapters \cite{chen2022adaptformer, chen2022vision, hu2021lora}. In this work, we select LoRA \cite{hu2021lora} to adapt the selected FM due to its promising results reported in the literature, namely in biometrics \cite{chettaoui2024froundation}. LoRA employs a low-dimensional reparametrization strategy, demonstrating effectiveness comparable to training the full parameter space \cite{aghajanyan2020intrinsic} while greatly reducing the number of parameters that need to be updated. In this method, the pre-trained weights of the FM, $W_0 \in \mathbb{R}^{d \times k}$, are kept unchanged, and trainable rank-decomposition matrices are added within each layer of the transformer, enabling effective adaptation with minimal computational overhead. The low-rank decomposition introduced by LoRA updates $W_0$ as follows:
\vspace{-1mm}
\begin{equation}
    W_0+ \Delta W=W_0+\gamma_r BA,
\end{equation} 
where $B \in \mathbb{R}^{d \times r}$ and $A \in \mathbb{R}^{r \times k}$ are the trainable rank-decomposition matrices, with the rank $r<<min(d,k)$, and $\gamma_r$ is a scaling factor. Originally, $\gamma_r$ was defined as $\frac{\alpha}{r}$, but this formulation often causes gradient collapse as the rank $r$ increases, resulting in a lack of performance gains despite the use of additional trainable parameters for fine-tuning \cite{kalajdzievski2023rank}. To address this issue, rank-stabilized LoRA (rsLoRA) \cite{kalajdzievski2023rank} modifies the scaling factor to $\frac{\alpha}{\sqrt{r}}$, preventing gradient collapse and enabling better performance at higher ranks. For this reason, we opt to employ rsLoRA to fine-tune CLIP, setting $\gamma_r = \frac{\alpha}{\sqrt{r}}$. Since $\gamma_r$ is a constant and the original CLIP weights are kept frozen, only the matrices $A$ and $B$ are updated. Once the adaption process is complete, the final model weights are calculated as $W = W_0 + \gamma_r BA$. As no extra parameters are added, the original model's computational efficiency during inference is maintained.

The FM's image encoder contains alternating multiheaded self-attention (MSA) layers and multilayer perceptron (MLP) blocks. Layer normalization and residual connections are applied before and after each block, respectively. For simplicity and parameter efficiency, LoRA is only applied on the MSA weights, leaving the MLP unaltered \cite{hu2021lora}. Although LoRA can be applied to the query, key, value and output ($q$, $k$, $v$ and $o$, respectively) projection matrices in the MSA, we only adapt $q$ and $v$ matrices following the results obtained in the original study where LoRA was proposed \cite{hu2021lora} and recent work using FMs in FR \cite{chettaoui2024froundation}. The MSA layers run $h$ parallel heads, each with a unique set of $q$, $k$ and $v$. The LoRA layers in each head function independently and have distinct weights. When an embedding $x$ is fed to the MSA, the $q$, $k$ and $v$ projection layers in head $i$ ($Q_i$, $K_i$ and $V_i$, respectively) are calculated as follows:
\vspace{-1mm}
\begin{equation}
    \label{eq:qkv_projection}
    \begin{gathered}
        Q_i=W_i^qx+\gamma_r B_i^qA_i^qx,\\
        K_i=W_i^kx,\\
        V_i=W_i^vx+\gamma_r B_i^vA_i^vx,
    \end{gathered}
\end{equation}
\vspace{-1mm}
where $W_i^q$, $W_i^k$ and $W_i^v$ are the frozen projection layers for $q$, $k$ and $v$, respectively, and $A_i^q$, $B_i^q$, $A_i^v$ and $B_i^v$ correspond to the trainable LoRA layers. $Q_i$, $K_i$ and $V_i$ are then used to compute the attention score of head $i$, using the dimension of the key vectors, $d_k$, as a scaling factor: 

\vspace{-2mm}
\begin{equation}
    \label{eq:attention}
    Attention(Q_i,K_i,V_i)=Softmax\Big(\frac{Q_iK_i^T}{\sqrt{d_k}}\Big)V_i.
\end{equation}
\vspace{-2mm}

The MSA layer's output is determined by concatenating all heads' attention scores along the feature dimension and feeding the resultant vector to the projection layer $O$:
\vspace{-1mm}
\begin{equation}
    \label{eq:multihead}
    Multihead(Q,V,K)=Concat(head_1, ..., head_k)W^0.
\end{equation}
\vspace{-1mm}
The final output of the MSA is processed by the frozen MLP, completing the execution of a ViT block. The output of block $l$ is then processed by block $l+1$, which consists of a new MSA adapted with LoRA followed by a frozen MLP. 

\textbf{Classification:} Using LoRA to fine-tune CLIP allows it to produce a final embedding space adapted to the downstream MAD task. In this scenario, the FM works as a feature extractor on top of which classification can be performed. Hence, an additional fully connected layer with two output neurons followed by a softmax layer is added on top of the FM, resulting in the complete detector architecture. This layer is trained along with fine-tuning the LoRA parameters, using the binary cross-entropy loss:
\vspace{-1mm}
\begin{equation}
    \label{eq:bce}
    L_{BCE}=-(y\, log(\tilde{y})+(1-y)\,log(1-\tilde{y})),
\end{equation}
where $y$ and $\tilde{y}$ represent the sample's ground truth and predicted labels, respectively. After training, and during the feedforward MAD process, all MADation's weights are kept frozen. The detection score is obtained from the output of the binary classification layer, with the highest output score defining the model prediction for each sample.
\vspace{-1mm}
\subsection{MAD Baselines} \label{sec:baseline}
\vspace{-2mm}
To provide a comprehensive analysis of the use of FMs to perform MAD and prove the effectiveness of MADation when compared with alternative baseline solutions, we considered three alternative FM or transformer-based approaches for MAD.

\textbf{Text-Image (TI) MAD:} FMs like CLIP have demonstrated exceptional zero-shot learning performance across several downstream tasks, including action recognition in videos, sentiment analysis and car model classification \cite{radford2021learning}. Taking this into account, we evaluate the selected FM zero-shot learning performance on MAD, by simultaneously using its text and image encoders. To this end, TI processes image-text pairs where the textual input specifies the two possible classification labels. The predicted label is determined by analysing the similarity score between the image embedding and the input text embeddings. However, the potential of this technique should not be overestimated based on its success in general downstream domains such as the ones specified above given the domain-specific nature of MAD. The fact that no adaption is performed to the specific requirements of the downstream MAD task makes TI prone to underperform when compared with methods that consider MAD's specific characteristics, such as MADation. 

\textbf{ViT Trained from Scratch (ViT-FS) MAD:} The remarkable performance of FMs is closely related to their underlying architectures, which are often based on ViT networks \cite{alexey2020image}. These architectures have demonstrated promising performance in tasks such as MAD \cite{zhang2024generalized} and presentation attack detection (PAD) \cite{huang2022adaptive}. In this work, we also examine whether ViT networks can effectively perform MAD, allowing us to assess their contribution to the proposed FM-based methodology, MADation. Specifically, we train from scratch the same ViT-B and ViT-L used to evaluate MADation, using only the selected MAD training datasets. These transformers' parameters are randomly initialized, meaning that ViT-FS does not benefit from prior knowledge acquired during massive training and thus cannot be considered an FM. This approach enables a direct comparison between FM-based methods, such as MADation, and visual transformers, allowing us to assess how valuable the in-built knowledge of FMs is for downstream tasks such as MAD.

\textbf{Feature Extractor (FE) MAD:} To determine the importance of the network adaption allowed by LoRA in MADation, we develop an experiment where the FM is not fine-tuned. In this scenario, the FM works as a frozen feature extractor on top of which a fully connected layer is trained to perform classification, using the binary cross-entropy loss (Equation \ref{eq:bce}). This experiment allows us to assess the suitability of the FM's original feature space to discriminate between MAD classes while quantitatively measuring the improvements introduced by adapting the FM's weights to the domain-specific downstream MAD task with LoRA.

\vspace{-3mm}
\section{Experimental Setup} \label{sec:exp_setup}
\vspace{-2mm}

\textbf{Datasets:} The Synthetic Morphing Attack Detection Development (SMDD) dataset \cite{SMDD} was selected as the training dataset of the proposed models. SMDD \cite{SMDD} is a synthetic-based MAD dataset, consisting of 25k bona-fide images generated using the StyleGAN2-ADA framework \cite{DBLP:conf/nips/KarrasAHLLA20, DBLP:conf/iwbf/VenkateshZRRDB20} and 15k morphing attacks created from the bona-fide samples using the OpenCV morphing technique \cite{openCVmorph}. The SMDD dataset was chosen for training as it ensures privacy by avoiding the use of real face images and because it has shown remarkable success in MAD solution development \cite{damer2023mordiff, neto2022orthomad, caldeira2023unveiling} and public competitions \cite{huber2022syn}. The benchmarking datasets proposed in these works, MAD22 \cite{huber2022syn} and its extension MorDIFF \cite{damer2023mordiff}, were also chosen to ensure the results' comparability and a data domain identity disjoint from the SMDD training data. The evaluation benchmarks are based on the Face Research Lab London (FRLL) dataset \cite{DeBruine2021} and thus contain the same 204 bona-fide images. Additionally, the same image pairs were used to create the morphing attacks in MAD22 and MorDIFF. The MAD22 dataset includes morphed images from five different approaches: three image-level techniques (FaceMorpher, OpenCV \cite{openCVmorph}, and Webmorph) and two GAN-based representation-level methods (MIPGAN I and II \cite{DBLP:journals/tbbis/ZhangVRRDB21}). The morphing samples of the MorDIFF dataset were generated with a diffusion autoencoder \cite{DBLP:conf/cvpr/PreechakulCWS22}. In addition to MAD22 \cite{huber2022syn}, the FRLL-Morphs dataset \cite{DBLP:journals/corr/abs-2012-05344} is utilized for evaluation so that a comparison is possible with methods that are not evaluated on MAD22. The FRLL-Morphs dataset \cite{DBLP:journals/corr/abs-2012-05344}, similarly derived from the Face Research London Lab dataset \cite{DeBruine2021}, serves as a benchmark for MAD evaluation. The dataset comprises 204 genuine samples and over 1,000 morphed faces per technique, generated using five distinct morphing methods: Style-GAN2 \cite{DBLP:conf/nips/KarrasAHLLA20, DBLP:conf/iwbf/VenkateshZRRDB20}, WebMorph \cite{debruine2018debruine}, AMSL \cite{neubert2018extended}, FaceMorpher \cite{quek2019facemorpher}, and OpenCV \cite{openCVmorph}.

\textbf{Image Pre-Processing:} Before being used as an input to the FM, each sample is cropped following \cite{SMDD} and then resized to $224\times224$ pixels to comply with the image resolution originally used to train CLIP \cite{radford2021learning}. During training, all samples are also subject to data augmentation using random horizontal flipping, following \cite{DBLP:journals/istr/SeiboldSHE20,neto2022orthomad}. Due to the success achieved by tokenization in FMs' NLP applications, Alexey~\textit{et al.}~\cite{alexey2020image} proposed preprocessing images into tokens before feeding them to the FM's image encoder. All the experiments mentioned in this document follow this tokenization process, represented in Figure \ref{fig:pipeline}. Each input sample is divided into non-overleaping regions which are fed to a linear projection layer \cite{alexey2020image}. The resultant patch embeddings are appended to a learnable embedding, the class (CLS) token \cite{alexey2020image}, which constitutes an image representation that helps classify the input into predefined categories. Additional position embeddings are considered to preserve the spatial order of the original sample's patches. The final embedding vector is then fed to the image encoder \cite{alexey2020image}.

\textbf{Model Architecture:} CLIP \cite{radford2021learning} released four different models with two architectures: base and large. CLIP's base architecture contains 86M parameters and has 2 variants with different patch sizes: 16 and 32. The large version of CLIP has 0.3 billion parameters and includes a variant pre-trained at a higher resolution of 336 pixels for one additional epoch to boost performance \cite{touvron2019fixing}. Following the results achieved by recent work that used CLIP in FR \cite{chettaoui2024froundation}, we decide to consider one version of each architecture, namely CLIP's base version with a patch size of 16 and CLIP's large version trained without high-resolution images, which we refer to as ViT-B and ViT-L, respectively. These architectures were used in all the settings described in Sections \ref{sec:MADation} and \ref{sec:baseline}, namely TI, ViT-FS, FE and MADation. 

\textbf{Implementation Details:} All models were trained for 40 epochs using the AdamW optimizer \cite{loshchilov2018decoupled} with momentum of 0.9 and weight decay of 0.05 \cite{chettaoui2024froundation}. For MADation, we used ViT-B and ViT-L architectures with model and header learning rates of 1e-5 and 1e-4, respectively. The LoRA parameters were set to $r=2$, $\alpha=\{4,8\}$ and $dropout=\{0.4,0.2\}$ for ViT-B and ViT-L respectively. For FE, we used a header learning rate of 1e-2 for both architectures. ViT-FS models were trained with learning rates of 1e-5/5e-5 (model/header) for ViT-B and 1e-4/1e-4 for ViT-L. The batch size was set to 256 for all settings.

\textbf{Evaluation Metrics:} The MAD evaluation metrics were selected to ensure compliance with the ISO/IEC 30107-3 \cite{ISO301073} standard and enable consistent benchmarking and comparability with previous studies \cite{huber2022syn, damer2023mordiff}. Performance is reported using the Bona-fide Presentation Classification Error Rate (BPCER), the Attack Presentation Classification Error Rate (APCER), and the detection Equal Error Rate (EER). The BPCER quantifies the proportion of bona-fide images misclassified as attack samples, while the APCER measures the proportion of attack images misclassified as bona-fide samples. The detection EER is the error rate at the operating point where the BPCER and APCER are equal, offering a concise measure of the system's overall performance balance. To cover different operational points and present comparative results, we report both the APCER at fixed BPCER values and the BPCER at fixed APCER values, evaluated at values of 1\%, 10\%, and 20\%.

\vspace{-2mm}
\section{Results and Discussion} \label{sec:results}
\vspace{-2mm}
\textbf{Zero-Shot MAD (TI):} As described in Section \ref{sec:baseline}, the FM's zero-shot performance was evaluated by simultaneously pairing images with textual prompts describing the two possible classification labels, `face image morphing attack' and `bona-fide presentation' \cite{ISO20059}. Table \ref{tab:our_methods} displays TI results for both ViT-B and ViT-L. The results obtained with ViT-B reveal the performance limitations of TI, as it performs close to random in 3 out of the 6 evaluated datasets and results in high EER values for all of them. Although a similar tendency is verified for ViT-L in some evaluation datasets, this network performs significantly better than ViT-B and even achieves competitive results with recent MAD solutions from the literature (Table \ref{tab:sota}) in MIPGAN I and MIPGAN II. The higher zero-shot MAD capacity of ViT-L is justified by its higher number of parameters, which allow it to learn a wider spectrum of features during its pre-training stage and thus perform better for a wider variety of tasks, as demonstrated in \cite{radford2021learning}. Nonetheless, the global performance of both networks in the TI scenario is still far from satisfactory in comparison to other options described later in this section, highlighting the limitations of FM's in domain-specific scenarios such as MAD. These limitations can be largely overcome by adapting the FM to the downstream MAD task, as will be later shown in this section.

\begin{table}[tbh!]
\footnotesize
\centering
\caption{Evaluation results for CLIP ViT-B and ViT-L for four sets of experiments: TI, ViT-FS, FE and MADation. 
The best and second-best results achieved for each metric in each test dataset are highlighted in bold and underlined, respectively.}
\vspace{-3mm}
\resizebox{0.47\textwidth}{!}
{
\begin{tabular}{|c|c|c|c|ccc|ccc|}
\hline
\multicolumn{2}{|c|}{\multirow{2}{*}{Method}}                                    & \multirow{2}{*}{Test data} & \multirow{2}{*}{EER (\%)} & \multicolumn{3}{c|}{APCER (\%) @ BPCER (\%)}                                                                          & \multicolumn{3}{c|}{BPCER (\%) @ APCER (\%)}                                                                          \\ \cline{5-10} 
\multicolumn{2}{|c|}{}                                                           &                            &                           & \multicolumn{1}{c|}{1.00}                    & \multicolumn{1}{c|}{10.00}                   & 20.00                   & \multicolumn{1}{c|}{1.00}                    & \multicolumn{1}{c|}{10.00}                   & 20.00                   \\ \hline
\multicolumn{1}{|c|}{\multirow{32}{*}{ViT-B}} & \multirow{8}{*}{TI}              & FaceMorph                  & 51.50                     & \multicolumn{1}{c|}{98.40}                   & \multicolumn{1}{c|}{88.20}                   & 81.40                   & \multicolumn{1}{c|}{99.51}                   & \multicolumn{1}{c|}{93.63}                   & 85.29                   \\
\multicolumn{1}{|c|}{}                        &                                  & MIPGAN\_I                  & 36.40                     & \multicolumn{1}{c|}{99.80}                   & \multicolumn{1}{c|}{81.10}                   & 65.30                   & \multicolumn{1}{c|}{86.76}                   & \multicolumn{1}{c|}{55.88}                   & 46.57                   \\
\multicolumn{1}{|c|}{}                        &                                  & MIPGAN\_II                 & 33.40                     & \multicolumn{1}{c|}{99.60}                   & \multicolumn{1}{c|}{76.00}                   & 55.30                   & \multicolumn{1}{c|}{80.39}                   & \multicolumn{1}{c|}{49.02}                   & 43.63                   \\
\multicolumn{1}{|c|}{}                        &                                  & OpenCV                     & 47.15                     & \multicolumn{1}{c|}{99.90}                   & \multicolumn{1}{c|}{83.74}                   & 74.90                   & \multicolumn{1}{c|}{98.04}                   & \multicolumn{1}{c|}{81.37}                   & 70.10                   \\
\multicolumn{1}{|c|}{}                        &                                  & WebMorph                   & 35.60                     & \multicolumn{1}{c|}{98.20}                   & \multicolumn{1}{c|}{70.20}                   & 57.20                   & \multicolumn{1}{c|}{86.76}                   & \multicolumn{1}{c|}{61.27}                   & 48.53                   \\
\multicolumn{1}{|c|}{}                        &                                  & MorDIFF                    & 51.90                     & \multicolumn{1}{c|}{100.00}                  & \multicolumn{1}{c|}{92.60}                   & 86.70                   & \multicolumn{1}{c|}{99.02}                   & \multicolumn{1}{c|}{92.65}                   & 85.29                   \\ \cline{3-10} 
\multicolumn{1}{|c|}{}                        &                                  & \textit{Average}           & \textit{42.66}            & \multicolumn{1}{c|}{\textit{99.32}}          & \multicolumn{1}{c|}{\textit{81.97}}          & \textit{70.13}          & \multicolumn{1}{c|}{\textit{91.75}}          & \multicolumn{1}{c|}{\textit{72.30}}          & \textit{63.24}          \\ \cline{3-10} 
\multicolumn{1}{|c|}{}                        &                                  & \textit{Worst}             & \textit{51.90}            & \multicolumn{1}{c|}{\textit{100.00}}         & \multicolumn{1}{c|}{\textit{92.60}}          & \textit{86.70}          & \multicolumn{1}{c|}{\textit{99.51}}          & \multicolumn{1}{c|}{\textit{93.63}}          & \textit{85.29}          \\ \cline{2-10} 
\multicolumn{1}{|c|}{}                        & \multirow{8}{*}{ViT-FS}          & FaceMorph                  & 5.38                      & \multicolumn{1}{c|}{8.77}                    & \multicolumn{1}{c|}{2.49}                    & 0.90                    & \multicolumn{1}{c|}{20.98}                   & \multicolumn{1}{c|}{{\ul 0.49}}              & \textbf{0.00}           \\
\multicolumn{1}{|c|}{}                        &                                  & MIPGAN\_I                  & 32.87                     & \multicolumn{1}{c|}{85.66}                   & \multicolumn{1}{c|}{61.35}                   & 47.41                   & \multicolumn{1}{c|}{100.00}                  & \multicolumn{1}{c|}{49.02}                   & 49.02                   \\
\multicolumn{1}{|c|}{}                        &                                  & MIPGAN\_II                 & 27.19                     & \multicolumn{1}{c|}{94.92}                   & \multicolumn{1}{c|}{64.94}                   & 44.42                   & \multicolumn{1}{c|}{100.00}                  & \multicolumn{1}{c|}{57.84}                   & 30.88                   \\
\multicolumn{1}{|c|}{}                        &                                  & OpenCV                     & 16.30                     & \multicolumn{1}{c|}{50.40}                   & \multicolumn{1}{c|}{26.42}                   & 14.27                   & \multicolumn{1}{c|}{100.00}                  & \multicolumn{1}{c|}{56.31}                   & 34.47                   \\
\multicolumn{1}{|c|}{}                        &                                  & WebMorph                   & 22.80                     & \multicolumn{1}{c|}{83.60}                   & \multicolumn{1}{c|}{58.00}                   & 44.40                   & \multicolumn{1}{c|}{100.00}                  & \multicolumn{1}{c|}{52.94}                   & 32.35                   \\
\multicolumn{1}{|c|}{}                        &                                  & MorDIFF                    & 28.14                     & \multicolumn{1}{c|}{84.73}                   & \multicolumn{1}{c|}{52.00}                   & 35.93                   & \multicolumn{1}{c|}{100.00}                  & \multicolumn{1}{c|}{56.31}                   & 34.37                   \\ \cline{3-10} 
\multicolumn{1}{|c|}{}                        &                                  & \textit{Average}           & \textit{22.13}            & \multicolumn{1}{c|}{\textit{68.01}}          & \multicolumn{1}{c|}{\textit{44.20}}          & \textit{31.22}          & \multicolumn{1}{c|}{\textit{86.83}}          & \multicolumn{1}{c|}{\textit{40.68}}          & \textit{26.41}          \\ \cline{3-10} 
\multicolumn{1}{|c|}{}                        &                                  & \textit{Worst}             & \textit{32.87}            & \multicolumn{1}{c|}{\textit{94.92}}          & \multicolumn{1}{c|}{\textit{64.94}}          & \textit{47.41}          & \multicolumn{1}{c|}{\textit{100.00}}         & \multicolumn{1}{c|}{\textit{57.84}}          & \textit{49.02}          \\ \cline{2-10} 
\multicolumn{1}{|c|}{}                        & \multirow{8}{*}{FE}              & FaceMorph                  & 2.89                      & \multicolumn{1}{c|}{4.89}                    & \multicolumn{1}{c|}{{\ul 1.30}}              & {\ul 0.20}              & \multicolumn{1}{c|}{11.22}                   & \multicolumn{1}{c|}{{\ul 0.49}}              & {\ul 0.49}              \\
\multicolumn{1}{|c|}{}                        &                                  & MIPGAN\_I                  & 26.00                     & \multicolumn{1}{c|}{83.27}                   & \multicolumn{1}{c|}{55.68}                   & 36.06                   & \multicolumn{1}{c|}{77.94}                   & \multicolumn{1}{c|}{50.98}                   & 32.84                   \\
\multicolumn{1}{|c|}{}                        &                                  & MIPGAN\_II                 & 34.26                     & \multicolumn{1}{c|}{91.43}                   & \multicolumn{1}{c|}{74.70}                   & 57.27                   & \multicolumn{1}{c|}{84.80}                   & \multicolumn{1}{c|}{65.20}                   & 51.96                   \\
\multicolumn{1}{|c|}{}                        &                                  & OpenCV                     & 14.88                     & \multicolumn{1}{c|}{39.98}                   & \multicolumn{1}{c|}{20.34}                   & 9.21                    & \multicolumn{1}{c|}{61.27}                   & \multicolumn{1}{c|}{18.63}                   & 10.78                   \\
\multicolumn{1}{|c|}{}                        &                                  & WebMorph                   & 32.80                     & \multicolumn{1}{c|}{91.40}                   & \multicolumn{1}{c|}{71.40}                   & 49.80                   & \multicolumn{1}{c|}{84.80}                   & \multicolumn{1}{c|}{66.18}                   & 52.94                   \\
\multicolumn{1}{|c|}{}                        &                                  & MorDIFF                    & {\ul 17.86}               & \multicolumn{1}{c|}{50.90}                   & \multicolumn{1}{c|}{27.05}                   & {\ul 13.77}             & \multicolumn{1}{c|}{{\ul 59.22}}             & \multicolumn{1}{c|}{{\ul 24.27}}             & {\ul 12.62}             \\ \cline{3-10} 
\multicolumn{1}{|c|}{}                        &                                  & \textit{Average}           & \textit{21.45}            & \multicolumn{1}{c|}{\textit{60.31}}          & \multicolumn{1}{c|}{\textit{41.74}}          & \textit{27.72}          & \multicolumn{1}{c|}{\textit{63.21}}          & \multicolumn{1}{c|}{\textit{37.62}}          & \textit{26.94}          \\ \cline{3-10} 
\multicolumn{1}{|c|}{}                        &                                  & \textit{Worst}             & \textit{34.26}            & \multicolumn{1}{c|}{\textit{91.43}}          & \multicolumn{1}{c|}{\textit{74.70}}          & \textit{57.27}          & \multicolumn{1}{c|}{{\ul \textit{84.80}}}    & \multicolumn{1}{c|}{\textit{66.18}}          & \textit{52.94}          \\ \cline{2-10} 
\multicolumn{1}{|c|}{}                        & \multirow{8}{*}{MADation (ours)} & FaceMorph                  & \textbf{0.00}             & \multicolumn{1}{c|}{\textbf{0.00}}           & \multicolumn{1}{c|}{\textbf{0.00}}           & \textbf{0.00}           & \multicolumn{1}{c|}{\textbf{0.00}}           & \multicolumn{1}{c|}{\textbf{0.00}}           & \textbf{0.00}           \\
\multicolumn{1}{|c|}{}                        &                                  & MIPGAN\_I                  & 33.37                     & \multicolumn{1}{c|}{82.97}                   & \multicolumn{1}{c|}{55.18}                   & 43.92                   & \multicolumn{1}{c|}{94.12}                   & \multicolumn{1}{c|}{72.55}                   & 52.94                   \\
\multicolumn{1}{|c|}{}                        &                                  & MIPGAN\_II                 & 22.21                     & \multicolumn{1}{c|}{79.98}                   & \multicolumn{1}{c|}{34.66}                   & 24.30                   & \multicolumn{1}{c|}{84.80}                   & \multicolumn{1}{c|}{47.55}                   & 26.47                   \\
\multicolumn{1}{|c|}{}                        &                                  & OpenCV                     & {\ul 3.85}                & \multicolumn{1}{c|}{{\ul 11.64}}             & \multicolumn{1}{c|}{{\ul 1.82}}              & {\ul 1.11}              & \multicolumn{1}{c|}{{\ul 23.53}}             & \multicolumn{1}{c|}{{\ul 0.98}}              & \textbf{0.00}           \\
\multicolumn{1}{|c|}{}                        &                                  & WebMorph                   & \textbf{10.80}            & \multicolumn{1}{c|}{{\ul 60.00}}             & \multicolumn{1}{c|}{\textbf{11.40}}          & \textbf{5.00}           & \multicolumn{1}{c|}{\textbf{51.47}}          & \multicolumn{1}{c|}{\textbf{11.76}}          & \textbf{4.41}           \\
\multicolumn{1}{|c|}{}                        &                                  & MorDIFF                    & \textbf{1.10}             & \multicolumn{1}{c|}{\textbf{1.60}}           & \multicolumn{1}{c|}{\textbf{0.00}}           & \textbf{0.00}           & \multicolumn{1}{c|}{\textbf{1.94}}           & \multicolumn{1}{c|}{\textbf{0.00}}           & \textbf{0.00}           \\ \cline{3-10} 
\multicolumn{1}{|c|}{}                        &                                  & \textit{Average}           & \textit{\textbf{11.89}}   & \multicolumn{1}{c|}{{\ul \textit{39.36}}}    & \multicolumn{1}{c|}{{\ul \textit{17.18}}}    & {\ul \textit{12.39}}    & \multicolumn{1}{c|}{\textit{\textbf{42.64}}} & \multicolumn{1}{c|}{{\ul \textit{22.14}}}    & {\ul \textit{13.97}}    \\ \cline{3-10} 
\multicolumn{1}{|c|}{}                        &                                  & \textit{Worst}             & \textit{33.37}            & \multicolumn{1}{c|}{{\ul \textit{82.97}}}    & \multicolumn{1}{c|}{{\ul \textit{55.18}}}    & \textit{43.92}          & \multicolumn{1}{c|}{\textit{94.12}}          & \multicolumn{1}{c|}{\textit{72.55}}          & \textit{52.94}          \\ \hline
\multicolumn{1}{|c|}{\multirow{32}{*}{ViT-L}} & \multirow{8}{*}{TI}              & FaceMorph                  & 44.60                     & \multicolumn{1}{c|}{98.40}                   & \multicolumn{1}{c|}{79.70}                   & 63.60                   & \multicolumn{1}{c|}{99.02}                   & \multicolumn{1}{c|}{87.25}                   & 76.96                   \\
\multicolumn{1}{|c|}{}                        &                                  & MIPGAN\_I                  & \textbf{18.90}            & \multicolumn{1}{c|}{{\ul 71.80}}             & \multicolumn{1}{c|}{{\ul 32.20}}             & \textbf{17.80}          & \multicolumn{1}{c|}{\textbf{69.61}}          & \multicolumn{1}{c|}{\textbf{33.82}}          & {\ul 18.14}             \\
\multicolumn{1}{|c|}{}                        &                                  & MIPGAN\_II                 & {\ul 12.80}               & \multicolumn{1}{c|}{{\ul 56.70}}             & \multicolumn{1}{c|}{{\ul 17.00}}             & {\ul 8.90}              & \multicolumn{1}{c|}{\textbf{59.31}}          & \multicolumn{1}{c|}{{\ul 17.16}}             & {\ul 8.33}              \\
\multicolumn{1}{|c|}{}                        &                                  & OpenCV                     & 35.47                     & \multicolumn{1}{c|}{96.24}                   & \multicolumn{1}{c|}{77.54}                   & 63.11                   & \multicolumn{1}{c|}{96.08}                   & \multicolumn{1}{c|}{73.53}                   & 55.39                   \\
\multicolumn{1}{|c|}{}                        &                                  & WebMorph                   & 25.20                     & \multicolumn{1}{c|}{94.80}                   & \multicolumn{1}{c|}{52.00}                   & 30.20                   & \multicolumn{1}{c|}{87.75}                   & \multicolumn{1}{c|}{50.98}                   & 32.35                   \\
\multicolumn{1}{|c|}{}                        &                                  & MorDIFF                    & 42.60                     & \multicolumn{1}{c|}{97.80}                   & \multicolumn{1}{c|}{79.60}                   & 69.50                   & \multicolumn{1}{c|}{97.06}                   & \multicolumn{1}{c|}{83.33}                   & 68.63                   \\ \cline{3-10} 
\multicolumn{1}{|c|}{}                        &                                  & \textit{Average}           & \textit{29.93}            & \multicolumn{1}{c|}{\textit{85.96}}          & \multicolumn{1}{c|}{\textit{56.34}}          & \textit{42.19}          & \multicolumn{1}{c|}{\textit{84.81}}          & \multicolumn{1}{c|}{\textit{57.68}}          & \textit{43.30}          \\ \cline{3-10} 
\multicolumn{1}{|c|}{}                        &                                  & \textit{Worst}             & \textit{44.60}            & \multicolumn{1}{c|}{\textit{98.40}}          & \multicolumn{1}{c|}{\textit{79.70}}          & \textit{69.50}          & \multicolumn{1}{c|}{\textit{99.02}}          & \multicolumn{1}{c|}{\textit{87.25}}          & \textit{76.96}          \\ \cline{2-10} 
\multicolumn{1}{|c|}{}                        & \multirow{8}{*}{ViT-FS}          & FaceMorph                  & 22.63                     & \multicolumn{1}{c|}{75.17}                   & \multicolumn{1}{c|}{38.68}                   & 24.93                   & \multicolumn{1}{c|}{88.29}                   & \multicolumn{1}{c|}{40.98}                   & 24.88                   \\
\multicolumn{1}{|c|}{}                        &                                  & MIPGAN\_I                  & 23.80                     & \multicolumn{1}{c|}{79.08}                   & \multicolumn{1}{c|}{42.93}                   & 25.50                   & \multicolumn{1}{c|}{91.18}                   & \multicolumn{1}{c|}{46.57}                   & 28.43                   \\
\multicolumn{1}{|c|}{}                        &                                  & MIPGAN\_II                 & 21.81                     & \multicolumn{1}{c|}{80.28}                   & \multicolumn{1}{c|}{36.65}                   & 25.40                   & \multicolumn{1}{c|}{91.67}                   & \multicolumn{1}{c|}{25.00}                   & 40.69                   \\
\multicolumn{1}{|c|}{}                        &                                  & OpenCV                     & 30.47                     & \multicolumn{1}{c|}{84.72}                   & \multicolumn{1}{c|}{59.92}                   & 44.23                   & \multicolumn{1}{c|}{94.12}                   & \multicolumn{1}{c|}{60.29}                   & 42.16                   \\
\multicolumn{1}{|c|}{}                        &                                  & WebMorph                   & 33.60                     & \multicolumn{1}{c|}{91.60}                   & \multicolumn{1}{c|}{59.80}                   & 48.60                   & \multicolumn{1}{c|}{100.00}                  & \multicolumn{1}{c|}{75.49}                   & 52.45                   \\
\multicolumn{1}{|c|}{}                        &                                  & MorDIFF                    & 40.92                     & \multicolumn{1}{c|}{94.51}                   & \multicolumn{1}{c|}{77.94}                   & 67.86                   & \multicolumn{1}{c|}{100.00}                  & \multicolumn{1}{c|}{81.55}                   & 67.96                   \\ \cline{3-10} 
\multicolumn{1}{|c|}{}                        &                                  & \textit{Average}           & \textit{28.87}            & \multicolumn{1}{c|}{\textit{84.23}}          & \multicolumn{1}{c|}{\textit{52.65}}          & \textit{39.42}          & \multicolumn{1}{c|}{\textit{94.21}}          & \multicolumn{1}{c|}{\textit{57.59}}          & \textit{40.15}          \\ \cline{3-10} 
\multicolumn{1}{|c|}{}                        &                                  & \textit{Worst}             & \textit{40.92}            & \multicolumn{1}{c|}{\textit{94.51}}          & \multicolumn{1}{c|}{\textit{77.94}}          & \textit{67.86}          & \multicolumn{1}{c|}{\textit{100.00}}         & \multicolumn{1}{c|}{\textit{81.55}}          & \textit{67.96}          \\ \cline{2-10} 
\multicolumn{1}{|c|}{}                        & \multirow{8}{*}{FE}              & FaceMorph                  & 9.77                      & \multicolumn{1}{c|}{44.17}                   & \multicolumn{1}{c|}{9.77}                    & 4.09                    & \multicolumn{1}{c|}{35.12}                   & \multicolumn{1}{c|}{10.24}                   & 5.37                    \\
\multicolumn{1}{|c|}{}                        &                                  & MIPGAN\_I                  & 23.51                     & \multicolumn{1}{c|}{88.84}                   & \multicolumn{1}{c|}{55.28}                   & 31.37                   & \multicolumn{1}{c|}{{\ul 71.57}}             & \multicolumn{1}{c|}{40.69}                   & 27.45                   \\
\multicolumn{1}{|c|}{}                        &                                  & MIPGAN\_II                 & 21.81                     & \multicolumn{1}{c|}{82.37}                   & \multicolumn{1}{c|}{45.42}                   & 25.10                   & \multicolumn{1}{c|}{{\ul 69.61}}             & \multicolumn{1}{c|}{32.84}                   & 23.53                   \\
\multicolumn{1}{|c|}{}                        &                                  & OpenCV                     & 15.89                     & \multicolumn{1}{c|}{55.77}                   & \multicolumn{1}{c|}{25.40}                   & 10.83                   & \multicolumn{1}{c|}{48.53}                   & \multicolumn{1}{c|}{22.06}                   & 12.75                   \\
\multicolumn{1}{|c|}{}                        &                                  & WebMorph                   & 26.40                     & \multicolumn{1}{c|}{86.60}                   & \multicolumn{1}{c|}{56.80}                   & 37.80                   & \multicolumn{1}{c|}{{\ul 68.63}}             & \multicolumn{1}{c|}{41.67}                   & 29.90                   \\
\multicolumn{1}{|c|}{}                        &                                  & MorDIFF                    & 22.85                     & \multicolumn{1}{c|}{87.03}                   & \multicolumn{1}{c|}{50.70}                   & 29.14                   & \multicolumn{1}{c|}{67.48}                   & \multicolumn{1}{c|}{35.92}                   & 24.27                   \\ \cline{3-10} 
\multicolumn{1}{|c|}{}                        &                                  & \textit{Average}           & \textit{20.04}            & \multicolumn{1}{c|}{\textit{74.13}}          & \multicolumn{1}{c|}{\textit{40.56}}          & \textit{23.06}          & \multicolumn{1}{c|}{{\ul \textit{60.16}}}    & \multicolumn{1}{c|}{\textit{30.57}}          & \textit{20.54}          \\ \cline{3-10} 
\multicolumn{1}{|c|}{}                        &                                  & \textit{Worst}             & {\ul \textit{26.40}}      & \multicolumn{1}{c|}{\textit{88.84}}          & \multicolumn{1}{c|}{\textit{56.80}}          & {\ul \textit{37.80}}    & \multicolumn{1}{c|}{\textit{\textbf{71.57}}} & \multicolumn{1}{c|}{{\ul \textit{41.67}}}    & {\ul \textit{29.90}}    \\ \cline{2-10} 
\multicolumn{1}{|c|}{}                        & \multirow{8}{*}{MADation (ours)} & FaceMorph                  & {\ul 0.40}                & \multicolumn{1}{c|}{{\ul 0.40}}              & \multicolumn{1}{c|}{\textbf{0.00}}           & \textbf{0.00}           & \multicolumn{1}{c|}{{\ul 0.49}}              & \multicolumn{1}{c|}{\textbf{0.00}}           & \textbf{0.00}           \\
\multicolumn{1}{|c|}{}                        &                                  & MIPGAN\_I                  & {\ul 20.32}               & \multicolumn{1}{c|}{\textbf{55.88}}          & \multicolumn{1}{c|}{\textbf{29.08}}          & {\ul 20.32}             & \multicolumn{1}{c|}{79.41}                   & \multicolumn{1}{c|}{{\ul 35.78}}             & \textbf{15.69}          \\
\multicolumn{1}{|c|}{}                        &                                  & MIPGAN\_II                 & \textbf{9.06}             & \multicolumn{1}{c|}{\textbf{19.42}}          & \multicolumn{1}{c|}{\textbf{9.06}}           & \textbf{5.58}           & \multicolumn{1}{c|}{100.00}                  & \multicolumn{1}{c|}{\textbf{5.39}}           & \textbf{0.98}           \\
\multicolumn{1}{|c|}{}                        &                                  & OpenCV                     & \textbf{2.23}             & \multicolumn{1}{c|}{\textbf{3.74}}           & \multicolumn{1}{c|}{\textbf{1.32}}           & \textbf{0.71}           & \multicolumn{1}{c|}{\textbf{15.69}}          & \multicolumn{1}{c|}{\textbf{0.00}}           & \textbf{0.00}           \\
\multicolumn{1}{|c|}{}                        &                                  & WebMorph                   & {\ul 20.40}               & \multicolumn{1}{c|}{\textbf{47.60}}          & \multicolumn{1}{c|}{{\ul 20.40}}             & {\ul 20.40}             & \multicolumn{1}{c|}{82.35}                   & \multicolumn{1}{c|}{{\ul 37.25}}             & {\ul 13.24}             \\
\multicolumn{1}{|c|}{}                        &                                  & MorDIFF                    & 19.26                     & \multicolumn{1}{c|}{{\ul 48.40}}             & \multicolumn{1}{c|}{{\ul 24.45}}             & 19.26                   & \multicolumn{1}{c|}{84.47}                   & \multicolumn{1}{c|}{34.95}                   & 15.53                   \\ \cline{3-10} 
\multicolumn{1}{|c|}{}                        &                                  & \textit{Average}           & {\ul \textit{11.94}}      & \multicolumn{1}{c|}{\textit{\textbf{29.24}}} & \multicolumn{1}{c|}{\textit{\textbf{14.05}}} & \textit{\textbf{11.04}} & \multicolumn{1}{c|}{\textit{60.40}}          & \multicolumn{1}{c|}{\textit{\textbf{18.90}}} & \textit{\textbf{7.57}}  \\ \cline{3-10} 
\multicolumn{1}{|c|}{}                        &                                  & \textit{Worst}             & \textit{\textbf{20.40}}   & \multicolumn{1}{c|}{\textit{\textbf{55.88}}} & \multicolumn{1}{c|}{\textit{\textbf{29.08}}} & {\ul \textit{20.40}}    & \multicolumn{1}{c|}{\textit{100.00}}         & \multicolumn{1}{c|}{\textit{\textbf{37.25}}} & \textit{\textbf{15.69}} \\ \hline
\end{tabular}}
\vspace{-4mm}
\label{tab:our_methods}
\end{table}

\begin{table}[tbh!]
\footnotesize
\centering
\caption{Results comparison between MADation and previous MAD solutions. All methods are trained on SMDD \cite{SMDD}, and evaluated on MAD22 \cite{huber2022syn} and its extension MorDIFF \cite{damer2023mordiff}. Specific values unavailable in the original papers are marked with ``-''. The best and second-best results achieved for each metric in each test dataset are highlighted in bold and underlined, respectively.}
\vspace{-3mm}
\resizebox{0.47\textwidth}{!}{
\begin{tabular}{|cc|c|c|ccc|ccc|}
\hline
\multicolumn{2}{|c|}{\multirow{2}{*}{Method}} & \multirow{2}{*}{Test data} & \multirow{2}{*}{EER (\%)} & \multicolumn{3}{c|}{APCER (\%) @ BPCER (\%)} & \multicolumn{3}{c|}{BPCER (\%) @ APCER (\%)}                      \\ \cline{5-10}
&                            &                &           & \multicolumn{1}{c|}{1.00}   & \multicolumn{1}{c|}{10.00}  & 20.00 & \multicolumn{1}{c|}{1.00}   & \multicolumn{1}{c|}{10.00}  & 20.00  \\ \hline
\multicolumn{2}{|c|}{\multirow{6}{*}{MixFaceNet-MAD \cite{SMDD, damer2023mordiff}}}                         & FaceMorph                  & 4.60                      & \multicolumn{1}{c|}{5.50}   & \multicolumn{1}{c|}{\underline{3.60}}   & 2.90 & \multicolumn{1}{c|}{-}  & \multicolumn{1}{c|}{-}   & -  \\ 
     && MIPGAN\_I                  & 16.70                     & \multicolumn{1}{c|}{75.80}  & \multicolumn{1}{c|}{22.20}  & 14.50 & \multicolumn{1}{c|}{-}  & \multicolumn{1}{c|}{-}   & - \\
     &&  MIPGAN\_II                 & 20.62                     & \multicolumn{1}{c|}{81.58}  & \multicolumn{1}{c|}{32.03}  & 20.62 & \multicolumn{1}{c|}{-}  & \multicolumn{1}{c|}{-}   & - \\
     &&  OpenCV                     & 8.33                      & \multicolumn{1}{c|}{36.38}  & \multicolumn{1}{c|}{6.50}   & 3.76 & \multicolumn{1}{c|}{-}  & \multicolumn{1}{c|}{-}   & - \\ 
     &&  WebMorph                   & 18.20                     & \multicolumn{1}{c|}{74.00}  & \multicolumn{1}{c|}{24.00}  & 17.60 & \multicolumn{1}{c|}{-}  & \multicolumn{1}{c|}{-}   & - \\ 
     && MorDIFF             & 8.50                     & \multicolumn{1}{c|}{33.40}  & \multicolumn{1}{c|}{7.40}   & 4.10  & \multicolumn{1}{c|}{-}  & \multicolumn{1}{c|}{-}   & -\\ \hline
\multicolumn{2}{|c|}{\multirow{6}{*}{Inception-MAD \cite{ramachandra2019detecting, damer2023mordiff}}}              & FaceMorph                  & \textbf{0.00}                      & \multicolumn{1}{c|}{1.70}   & \multicolumn{1}{c|}{\textbf{0.00}}   & \textbf{0.00} & \multicolumn{1}{c|}{-}  & \multicolumn{1}{c|}{-}   & - \\  
    &&  MIPGAN\_I                  & \underline{10.90}                     & \multicolumn{1}{c|}{\underline{50.90}}  & \multicolumn{1}{c|}{\textbf{13.70}}  & 5.70  & \multicolumn{1}{c|}{-}  & \multicolumn{1}{c|}{-}   & - \\ 
   &&  MIPGAN\_II                 & 16.22                     & \multicolumn{1}{c|}{82.48}  & \multicolumn{1}{c|}{25.83}  & 11.41 & \multicolumn{1}{c|}{-}  & \multicolumn{1}{c|}{-}   & - \\ 
    &&  OpenCV                     & 7.52                      & \multicolumn{1}{c|}{28.66}  & \multicolumn{1}{c|}{5.49}   & 3.05 & \multicolumn{1}{c|}{-}  & \multicolumn{1}{c|}{-}   & -\\ 
    &&  WebMorph                   & 18.00                     & \multicolumn{1}{c|}{85.20}  & \multicolumn{1}{c|}{27.40}  & 13.40 & \multicolumn{1}{c|}{-}  & \multicolumn{1}{c|}{-}   & - \\
    &&  MorDIFF             & \underline{5.30}                      & \multicolumn{1}{c|}{\underline{17.20}}  & \multicolumn{1}{c|}{\underline{3.50}}   & \underline{2.50}  & \multicolumn{1}{c|}{-}  & \multicolumn{1}{c|}{-}   & -\\  \hline
\multicolumn{2}{|c|}{\multirow{6}{*}{MorphHRNet \cite{DBLP:journals/corr/abs-2207-00899, huber2022syn}}}              & FaceMorph                  &          5.90             & \multicolumn{1}{c|}{31.20}   & \multicolumn{1}{c|}{4.30}   &  \underline{2.40} & \multicolumn{1}{c|}{48.04}  & \multicolumn{1}{c|}{1.96}  & \underline{1.47} \\  
   &&  MIPGAN\_I                  &     15.30                & \multicolumn{1}{c|}{89.80}  & \multicolumn{1}{c|}{21.90}  & 13.00 & \multicolumn{1}{c|}{\underline{75.98}}  & \multicolumn{1}{c|}{\underline{24.02}} & \underline{11.27} \\ 
    &&  MIPGAN\_II                 &   10.41                   & \multicolumn{1}{c|}{84.18}  & \multicolumn{1}{c|}{\underline{11.01}}  & 6.11 & \multicolumn{1}{c|}{61.27}  & \multicolumn{1}{c|}{\underline{11.27}}   & \underline{2.94} \\ 
    &&  OpenCV                     &        5.69             & \multicolumn{1}{c|}{66.97}  & \multicolumn{1}{c|}{\underline{3.76}}   & 1.63 & \multicolumn{1}{c|}{\underline{33.82}}  & \multicolumn{1}{c|}{\underline{1.96}}   & \underline{1.47} \\ 
  &&  WebMorph                   &   9.80                  & \multicolumn{1}{c|}{90.02}  & \multicolumn{1}{c|}{\underline{11.20}}  & \underline{4.20} & \multicolumn{1}{c|}{56.86}  & \multicolumn{1}{c|}{\underline{10.78}}   & \underline{3.92} \\
  &&  MorDIFF             &          -          & \multicolumn{1}{c|}{-}  & \multicolumn{1}{c|}{-}   & - & \multicolumn{1}{c|}{-}  & \multicolumn{1}{c|}{-}   & - \\  \hline 
\multicolumn{2}{|c|}{\multirow{6}{*}{Con-Text Net A \cite{huber2022syn}}}              & FaceMorph                  &      \textbf{0.00}                 & \multicolumn{1}{c|}{99.90}   & \multicolumn{1}{c|}{\textbf{0.00}}   &  \textbf{0.00} & \multicolumn{1}{c|}{100.00}  & \multicolumn{1}{c|}{\textbf{0.00}}   & \textbf{0.00} \\  
    &&  MIPGAN\_I                  &     12.30                & \multicolumn{1}{c|}{\textbf{41.90}}  & \multicolumn{1}{c|}{\underline{14.10}}  & 8.10 & \multicolumn{1}{c|}{\textbf{59.31}}  & \multicolumn{1}{c|}{\textbf{16.18}}   & \textbf{6.37} \\ 
     &&  MIPGAN\_II                 & 12.91                     & \multicolumn{1}{c|}{\underline{43.44}}  & \multicolumn{1}{c|}{14.51}  & 8.61 & \multicolumn{1}{c|}{\underline{59.31}}  & \multicolumn{1}{c|}{19.61}   & 5.88 \\ 
    &&  OpenCV                     &        17.48             & \multicolumn{1}{c|}{70.93}  & \multicolumn{1}{c|}{26.52}   & 15.75 & \multicolumn{1}{c|}{74.02}  & \multicolumn{1}{c|}{32.84}   & 16.18 \\ 
     &&  WebMorph                   &        26.20              & \multicolumn{1}{c|}{89.20}  & \multicolumn{1}{c|}{45.60}  & 31.00 & \multicolumn{1}{c|}{93.14}  & \multicolumn{1}{c|}{48.53}   & 31.86 \\
     &&  MorDIFF             &          -          & \multicolumn{1}{c|}{-}  & \multicolumn{1}{c|}{-}   & - & \multicolumn{1}{c|}{-}  & \multicolumn{1}{c|}{-}   & - \\  \hline 
\multicolumn{2}{|c|}{\multirow{6}{*}{E-CBAM@VCMI \cite{huber2022syn}}}              & FaceMorph                  &       41.20                & \multicolumn{1}{c|}{100.00}   & \multicolumn{1}{c|}{92.80}   &  62.80 & \multicolumn{1}{c|}{100.00}  & \multicolumn{1}{c|}{95.10}   & 82.35\\  
     &&  MIPGAN\_I                  &     32.50                & \multicolumn{1}{c|}{99.90}  & \multicolumn{1}{c|}{84.90}  & 60.20 & \multicolumn{1}{c|}{78.92}  & \multicolumn{1}{c|}{53.43}   &  40.69 \\ 
     &&  MIPGAN\_II                 &         25.93             & \multicolumn{1}{c|}{99.60}  & \multicolumn{1}{c|}{64.66}  & 37.34 & \multicolumn{1}{c|}{\textbf{56.86}}  & \multicolumn{1}{c|}{30.39}   & 30.39\\ 
     &&  OpenCV                     &            27.54         & \multicolumn{1}{c|}{98.58}  & \multicolumn{1}{c|}{48.68}   & 33.03 & \multicolumn{1}{c|}{76.47}  & \multicolumn{1}{c|}{50.49}   & 38.24 \\ 
     &&  WebMorph                   &           30.60           & \multicolumn{1}{c|}{99.00}  & \multicolumn{1}{c|}{86.80}  & 46.80 & \multicolumn{1}{c|}{69.12}  & \multicolumn{1}{c|}{47.06}   & 38.24\\
     &&  MorDIFF             &          -          & \multicolumn{1}{c|}{-}  & \multicolumn{1}{c|}{-}   & - & \multicolumn{1}{c|}{-}  & \multicolumn{1}{c|}{-}   & - \\  \hline 
\multicolumn{2}{|c|}{\multirow{6}{*}{Con-Text Net B \cite{huber2022syn}}}              & FaceMorph                  &    \textbf{0.00}                   & \multicolumn{1}{c|}{\textbf{0.00}}   & \multicolumn{1}{c|}{\textbf{0.00}}   &  \textbf{0.00} & \multicolumn{1}{c|}{\textbf{0.00}}  & \multicolumn{1}{c|}{\textbf{0.00}}   & \textbf{0.00} \\  
    &&  MIPGAN\_I                  &   30.30                  & \multicolumn{1}{c|}{71.50}  & \multicolumn{1}{c|}{53.00}  & 39.80 & \multicolumn{1}{c|}{87.75}  & \multicolumn{1}{c|}{60.29}   & 35.59  \\ 
     &&  MIPGAN\_II                 &   29.43                  & \multicolumn{1}{c|}{67.67}  & \multicolumn{1}{c|}{51.25}  & 39.14 & \multicolumn{1}{c|}{91.18}  & \multicolumn{1}{c|}{61.76}   & 47.06\\ 
     &&  OpenCV                     &          22.66           & \multicolumn{1}{c|}{57.83}  & \multicolumn{1}{c|}{34.45}   & 23.68 & \multicolumn{1}{c|}{82.35}  & \multicolumn{1}{c|}{43.14}   & 23.53 \\ 
     &&  WebMorph                   &           31.40           & \multicolumn{1}{c|}{81.00}  & \multicolumn{1}{c|}{59.60}  & 43.80 & \multicolumn{1}{c|}{94.61}  & \multicolumn{1}{c|}{55.39}   & 43.80 \\
     &&  MorDIFF             &          -          & \multicolumn{1}{c|}{-}  & \multicolumn{1}{c|}{-}   & - & \multicolumn{1}{c|}{-}  & \multicolumn{1}{c|}{-}   & - \\  \hline 
\multicolumn{2}{|c|}{\multirow{6}{*}{Xception \cite{DBLP:journals/corr/abs-2207-00899, huber2022syn}}}              & FaceMorph                  &           0.60            & \multicolumn{1}{c|}{0.50}   & \multicolumn{1}{c|}{\textbf{0.00}}   &  \textbf{0.00} & \multicolumn{1}{c|}{1.47}  & \multicolumn{1}{c|}{\underline{1.47}}   & \underline{1.47} \\  
     && MIPGAN\_I                  &     36.90               & \multicolumn{1}{c|}{97.90}  & \multicolumn{1}{c|}{80.40}  & 57.40 & \multicolumn{1}{c|}{86.27}  & \multicolumn{1}{c|}{57.35}   & 49.02 \\ 
     &&  MIPGAN\_II                 &      44.54              & \multicolumn{1}{c|}{99.50}  & \multicolumn{1}{c|}{92.49}  & 77.08 & \multicolumn{1}{c|}{90.20}  & \multicolumn{1}{c|}{67.65}   & 56.86 \\ 
     &&  OpenCV                     &            7.32           & \multicolumn{1}{c|}{\underline{21.75}}  & \multicolumn{1}{c|}{6.61}   &  2.54 & \multicolumn{1}{c|}{35.29}  & \multicolumn{1}{c|}{4.90}   & \underline{1.47} \\ 
     &&  WebMorph                   &        14.60              & \multicolumn{1}{c|}{\underline{49.40}}  & \multicolumn{1}{c|}{23.00}  & 10.80 & \multicolumn{1}{c|}{53.92}  & \multicolumn{1}{c|}{21.57}   & 11.76 \\
     &&  MorDIFF             &          -          & \multicolumn{1}{c|}{-}  & \multicolumn{1}{c|}{-}   & - & \multicolumn{1}{c|}{-}  & \multicolumn{1}{c|}{-}   & - \\  \hline 
\multicolumn{2}{|c|}{\multirow{6}{*}{D-FW-MixFaceNet \cite{DBLP:conf/icb/RachalwarFDD23}}}  & FaceMorph  &   \underline{0.10}   & \multicolumn{1}{c|}{-}   & \multicolumn{1}{c|}{-}   &  \textbf{0.00} & \multicolumn{1}{c|}{-}  & \multicolumn{1}{c|}{-} & - \\  
     && MIPGAN\_I   &          \textbf{6.70}  & \multicolumn{1}{c|}{-}   & \multicolumn{1}{c|}{-}   & \textbf{1.20}   & \multicolumn{1}{c|}{-}  & \multicolumn{1}{c|}{-} & -  \\ 
     &&  MIPGAN\_II    &           \underline{6.61}          & \multicolumn{1}{c|}{-}   & \multicolumn{1}{c|}{-}   &  \textbf{1.00} & \multicolumn{1}{c|}{-}  & \multicolumn{1}{c|}{-} & -  \\ 
     &&  OpenCV     &    13.72     & \multicolumn{1}{c|}{-}   & \multicolumn{1}{c|}{-}   &  9.04  & \multicolumn{1}{c|}{-}  & \multicolumn{1}{c|}{-} & -  \\ 
     &&  WebMorph      &   10.80    & \multicolumn{1}{c|}{-}   & \multicolumn{1}{c|}{-}   &  7.40  & \multicolumn{1}{c|}{-}  & \multicolumn{1}{c|}{-} & -  \\
     &&  MorDIFF             &          -          & \multicolumn{1}{c|}{-}  & \multicolumn{1}{c|}{-}   & - & \multicolumn{1}{c|}{-}  & \multicolumn{1}{c|}{-}   & - \\  \hline 
\multicolumn{2}{|c|}{\multirow{6}{*}{D-FW-CDCN \cite{DBLP:conf/icb/RachalwarFDD23}}}              & FaceMorph                  &     \textbf{0.00}          & \multicolumn{1}{c|}{}   & \multicolumn{1}{c|}{}   &  44.10 & \multicolumn{1}{c|}{}  & \multicolumn{1}{c|}{}   & \\  
     && MIPGAN\_I    &     11.90  & \multicolumn{1}{c|}{-}   & \multicolumn{1}{c|}{-}   & \underline{3.80}  & \multicolumn{1}{c|}{-}  & \multicolumn{1}{c|}{-} & -  \\
     &&  MIPGAN\_II   & 14.11   & \multicolumn{1}{c|}{-}   & \multicolumn{1}{c|}{-}   &  8.51  & \multicolumn{1}{c|}{-}  & \multicolumn{1}{c|}{-} & -  \\
     &&  OpenCV  &   \textbf{0.30}    & \multicolumn{1}{c|}{-}   & \multicolumn{1}{c|}{-}   &  \textbf{0.00}   & \multicolumn{1}{c|}{-}  & \multicolumn{1}{c|}{-} & -  \\
     &&  WebMorph    &   \textbf{0.00}   & \multicolumn{1}{c|}{-}   & \multicolumn{1}{c|}{-}   &  64.00   & \multicolumn{1}{c|}{-}  & \multicolumn{1}{c|}{-} & -  \\
     &&  MorDIFF             &          -          & \multicolumn{1}{c|}{-}  & \multicolumn{1}{c|}{-}   & - & \multicolumn{1}{c|}{-}  & \multicolumn{1}{c|}{-}   & - \\  \hline \hline
\multirow{12}{*}{MADation (ours)} 
& \multicolumn{1}{|c|}{\multirow{6}{*}{ViT-B}}  & FaceMorph      &        \textbf{0.00} 
             & \multicolumn{1}{c|}{\textbf{0.00}}   & \multicolumn{1}{c|}{\textbf{0.00}}   & \textbf{0.00} & \multicolumn{1}{c|}{\textbf{0.00}}  & \multicolumn{1}{c|}{\textbf{0.00}}   & \textbf{0.00} \\  
    & \multicolumn{1}{|c|}{} & MIPGAN\_I              &         33.37 
             & \multicolumn{1}{c|}{82.97}   & \multicolumn{1}{c|}{55.18}   & 43.92 & \multicolumn{1}{c|}{94.12}  & \multicolumn{1}{c|}{72.55}   & 52.94 \\ 
    & \multicolumn{1}{|c|}{} & MIPGAN\_II           &     22.21     
             & \multicolumn{1}{c|}{79.98}   & \multicolumn{1}{c|}{34.66}   & 24.30 & \multicolumn{1}{c|}{84.80}  & \multicolumn{1}{c|}{47.55}   & 26.47 \\ 
    & \multicolumn{1}{|c|}{} & OpenCV                    &          \underline{3.85}
             & \multicolumn{1}{c|}{\textbf{11.64}}   & \multicolumn{1}{c|}{\textbf{1.82}}   & \underline{1.11} & \multicolumn{1}{c|}{\textbf{23.53}}  & \multicolumn{1}{c|}{\textbf{0.98}}   &  \textbf{0.00} \\ 
    & \multicolumn{1}{|c|}{} & WebMorph        &        10.80 
             & \multicolumn{1}{c|}{60.00}   & \multicolumn{1}{c|}{11.40}   & 5.00 & \multicolumn{1}{c|}{\underline{51.47}}  & \multicolumn{1}{c|}{\textbf{11.76}}   & 4.41 \\
    & \multicolumn{1}{|c|}{} & MorDIFF    &        \textbf{1.10}  
             & \multicolumn{1}{c|}{\textbf{1.60}}   & \multicolumn{1}{c|}{\textbf{0.00}}   & \textbf{0.00} & \multicolumn{1}{c|}{\textbf{1.94}}  & \multicolumn{1}{c|}{\textbf{0.00}}   & \textbf{0.00} \\ \cline{2-10}
     & \multicolumn{1}{|c|}{\multirow{6}{*}{ViT-L}}   
                            & FaceMorph                  &         0.40             &  \multicolumn{1}{c|}{\underline{0.40}}   & \multicolumn{1}{c|}{\textbf{0.00}}   & \textbf{0.00}  & \multicolumn{1}{c|}{\underline{0.49}}  & \multicolumn{1}{c|}{\textbf{0.00}}   & \textbf{0.00} \\  
    & \multicolumn{1}{|c|}{} & MIPGAN\_I                  &           20.32           & \multicolumn{1}{c|}{55.88}  & \multicolumn{1}{c|}{29.08}  & 20.32 & \multicolumn{1}{c|}{79.41}  & \multicolumn{1}{c|}{35.78}   & 15.69 \\ 
    & \multicolumn{1}{|c|}{} & MIPGAN\_II                 &          9.06           & \multicolumn{1}{c|}{\textbf{19.42}}  & \multicolumn{1}{c|}{\textbf{9.06}}  & \underline{5.58} & \multicolumn{1}{c|}{100.00}  & \multicolumn{1}{c|}{\textbf{5.39}}   & \textbf{0.98} \\ 
    & \multicolumn{1}{|c|}{} & OpenCV                     &      19.26             & \multicolumn{1}{c|}{48.40}  & \multicolumn{1}{c|}{24.45}   & 19.26 & \multicolumn{1}{c|}{84.47}  & \multicolumn{1}{c|}{34.95}   & 15.53 \\ 
    & \multicolumn{1}{|c|}{} & WebMorph                   &         \underline{2.23}          & \multicolumn{1}{c|}{\textbf{3.74}}  & \multicolumn{1}{c|}{\textbf{1.32}}  & \textbf{0.71} & \multicolumn{1}{c|}{\textbf{15.69}}  & \multicolumn{1}{c|}{\textbf{0.00}}   &  \textbf{0.00}	\\
     & \multicolumn{1}{|c|}{} & MorDIFF             &         20.40           & \multicolumn{1}{c|}{47.60}  & \multicolumn{1}{c|}{20.40}   & 20.40 & \multicolumn{1}{c|}{\underline{82.35}}  & \multicolumn{1}{c|}{\underline{37.25}}   & \underline{13.24} \\ \hline
\end{tabular}}
\vspace{-5mm}
\label{tab:sota}
\end{table}

\textbf{Baselines toward MADation (ViT-FS and FE):} We further explore two alternative approaches, ViT-FS and FE. ViT-FS is trained from scratch, making it possible to assess the potential of the underlying ViT architectures for the MAD task without relying on the built-in knowledge of FMs. FE makes use of this knowledge by using CLIP as a frozen feature extractor on top of which a binary fully connected layer is trained to perform classification, allowing us to assess the suitability of the FM's original feature space to discriminate between MAD classes. The results achieved by both approaches are presented in Table \ref{tab:our_methods}. When the ViT-B architecture is considered, FE outperforms ViT-FS in 4 of the 6 considered benchmarking datasets, with an average EER difference of 0.68 pp.. For ViT-L, FE largely surpasses ViT-FS performance, improving the average EER by 8.83 pp.. Similar tendencies can be observed for metrics such as APCER@BPCER=20\% and APCER@BPCER=10\%. The superiority of FE when compared to ViT-FS possibly derives from the large number of trainable parameters of ViT-B and ViT-L. These large-scale networks require large amounts of training data to properly learn the considered task without overfitting, which might undermine ViT-FS' capacity to learn the MAD task given the reduced size of the SMDD dataset. On the other hand, FE benefits from the FM's previous knowledge, which was acquired during a pre-training phase with a massive amount of training data, which justifies its superior performance. Furthermore, the fact that this performance difference is higher when using ViT-L (8.83 pp. vs 0.68 pp. difference on average EER when using ViT-L and ViT-B, respectively) mostly derives from ViT-FS decreased performance in this scenario, which further reinforces that the limited capacity of ViT-FS to learn the MAD task is strongly correlated with the large number of trainable parameters of the considered networks. Hence, it can be concluded that taking advantage of the built-in knowledge of the FM results in better performance than training the network from scratch with a reduced amount of training data. Nonetheless, FE's MAD performance still has the potential for further enhancement, making it worth exploring whether adapting the FM to the downstream MAD task results in increased performance. 

\textbf{MADation:} As previously discussed, FMs can generalize to a wide variety of downstream tasks but can show limited capacity when handling domain-specific tasks such as MAD. The previously analysed FE results highlight this characteristic, since the EER values still leave space for improvement, revealing that the FM's feature space is most likely significantly misaligned for the MAD task. Although it is also possible that the selected classification network has saturated its capacity given the FM's feature space and should thus be deeper, we argue that the problem is most likely arising from the feature space misalignment which can be corrected through FM's adaption, as highlighted in previous studies \cite{zhang2024learning, chen2023adapting, chettaoui2024froundation}. Furthermore, this type of adaption allows the network to have more flexibility without giving up on the knowledge acquired during the FM's pre-training phase, which might constitute a good trade-off between the properties of FE and ViT-FS. Hence, we propose to adapt CLIP to the MAD task with LoRA layers, resulting in our proposed approach, MADation. We evaluate MADation using the same benchmarks as FE and ViT-FS, to allow for a fair comparison with these approaches, as shown in Table \ref{tab:our_methods}. It can be seen that MADation achieves the best and/or second-best performance levels for most of the evaluated metrics and benchmarks. In particular, ViT-B is the best-performing method on average in 2 out of the 7 evaluated metrics, and the second-best-performing method on the remaining 5. For these 5 metrics, ViT-L presents the best overall performance. The analysis of the average EER also reveals MADation's superiority when compared with the remaining approaches, as it surpasses ViT-FS and FE by 10.24 pp. and 9.56 pp., respectively, for ViT-B and by 16.93 pp. and 8.10 pp., respectively, for ViT-L. MADation's improvements when compared with FE prove the importance of performing a correct FM adaption to downstream domain-specific tasks such as MAD. Simultaneously, MADation's superiority regarding ViT-FS shows that the network adaption provided by LoRA does not prevent the final network from benefitting from the knowledge acquired during pre-training. Hence, it is possible to conclude that MADation reaches an efficient trade-off between preserving the FM's built-in knowledge and fine-tuning it to the downstream task, resulting in improved MAD performance.

\begin{table}[tbh!]
\footnotesize
\centering
\caption{Results comparison between MADation and previous MAD solutions. All methods are trained on SMDD \cite{SMDD}, and evaluated on FRLL-Morphs \cite{DBLP:journals/corr/abs-2012-05344}. Specific values unavailable in the original papers are marked with ``-''. The best and second-best results achieved for each metric in each test dataset are highlighted in bold and underlined, respectively.}
\vspace{-3mm}
\resizebox{0.47\textwidth}{!}{
\begin{tabular}{|cc|c|c|ccc|}
\hline
\multicolumn{2}{|c|}{\multirow{2}{*}{Method}} & \multirow{2}{*}{Test data} & \multirow{2}{*}{EER (\%)} & \multicolumn{3}{c|}{BPCER (\%) @ APCER (\%)}                      \\ \cline{5-7}
&                            &                &            & \multicolumn{1}{c|}{1.00}   & \multicolumn{1}{c|}{10.00}  & 20.00  \\ \hline
\multicolumn{2}{|c|}{\multirow{5}{*}{OrthoMAD \cite{neto2022orthomad}}}                         
                             & FRLL-Style-GAN2      & \underline{6.54}          & \multicolumn{1}{c|}{\underline{13.74}}    & \multicolumn{1}{c|}{-}      & \underline{3.76} \\
                             && FRLL-WebMorph       & 15.23                     & \multicolumn{1}{c|}{70.92}                & \multicolumn{1}{c|}{-}      & 9.50 \\
                             && FRLL-OpenCV         & \textbf{0.73}             & \multicolumn{1}{c|}{\textbf{0.73}}        & \multicolumn{1}{c|}{-}      & 0.32 \\
                             && FRLL-AMSL           & 14.80                     & \multicolumn{1}{c|}{65.05}                & \multicolumn{1}{c|}{-}      & 10.89 \\
                             && FRLL-FaceMorpher    & \underline{0.98}          & \multicolumn{1}{c|}{2.37}                 & \multicolumn{1}{c|}{-}      & \underline{0.08} \\ \hline         
     \multicolumn{2}{|c|}{\multirow{5}{*}{IDistill \cite{caldeira2023unveiling}}}                         
                             & FRLL-Style-GAN2      & \textbf{1.96}             & \multicolumn{1}{c|}{\textbf{8.51}}        & \multicolumn{1}{c|}{-}      & \textbf{0.08} \\
                             && FRLL-WebMorph       & \underline{4.01}          & \multicolumn{1}{c|}{14.41}                & \multicolumn{1}{c|}{-}      & \underline{0.33} \\
                             && FRLL-OpenCV         & 2.46                      & \multicolumn{1}{c|}{6.14}                 & \multicolumn{1}{c|}{-}      & \underline{0.16} \\
                             && FRLL-AMSL           & \underline{4.00}          & \multicolumn{1}{c|}{\underline{21.10}}    & \multicolumn{1}{c|}{-}      & \underline{2.85} \\
                             && FRLL-FaceMorpher    & 2.05                      & \multicolumn{1}{c|}{4.26}                 & \multicolumn{1}{c|}{-}      & 0.16 \\ \hline    
     \multicolumn{2}{|c|}{\multirow{5}{*}{MixFaceNet \cite{SMDD}}}                         
                             & FRLL-Style-GAN2      & 8.99          & \multicolumn{1}{c|}{42.16}  & \multicolumn{1}{c|}{\textbf{8.82}}          & 4.41 \\          
                             && FRLL-WebMorph       & 12.35         & \multicolumn{1}{c|}{80.39}  & \multicolumn{1}{c|}{15.20}                  & 7.84 \\
                             && FRLL-OpenCV         & 4.39          & \multicolumn{1}{c|}{26.47}  & \multicolumn{1}{c|}{1.96}                   & 1.47\\
                             && FRLL-AMSL           & 15.18         & \multicolumn{1}{c|}{49.51}  & \multicolumn{1}{c|}{21.08}                  & 11.76 \\
                             && FRLL-FaceMorpher    & 3.87          & \multicolumn{1}{c|}{23.53}  & \multicolumn{1}{c|}{0.49}                   & 0.49 \\ \hline
      \multicolumn{2}{|c|}{\multirow{5}{*}{PW-MAD \cite{SMDD}}}                         
                              & FRLL-Style-GAN2     & 16.64         & \multicolumn{1}{c|}{80.39}  & \multicolumn{1}{c|}{25.98}                  & 13.24 \\
                             && FRLL-WebMorph       & 16.65         & \multicolumn{1}{c|}{80.39}  & \multicolumn{1}{c|}{24.02}                  & 13.24 \\
                             && FRLL-OpenCV         & 2.42          & \multicolumn{1}{c|}{22.06}  & \multicolumn{1}{c|}{\underline{0.49}}       & 0.49\\
                             && FRLL-AMSL           & 15.18         & \multicolumn{1}{c|}{96.57}  & \multicolumn{1}{c|}{24.02}                  & 5.88 \\
                             && FRLL-FaceMorpher    & 2.20          & \multicolumn{1}{c|}{26.47}  & \multicolumn{1}{c|}{\underline{0.49}}       & \textbf{0.00} \\ \hline
      \multicolumn{2}{|c|}{\multirow{5}{*}{Inception \cite{SMDD}}}                         
                              & FRLL-Style-GAN2     & 11.37         & \multicolumn{1}{c|}{72.06}  & \multicolumn{1}{c|}{\underline{13.73}}      & 6.86 \\
                             && FRLL-WebMorph       & 9.86          & \multicolumn{1}{c|}{53.92}  & \multicolumn{1}{c|}{9.80}                   & 2.94 \\
                             && FRLL-OpenCV         & 5.38          & \multicolumn{1}{c|}{38.73}  & \multicolumn{1}{c|}{1.96}                   & 0.98 \\
                             && FRLL-AMSL           & 10.79         & \multicolumn{1}{c|}{72.06}  & \multicolumn{1}{c|}{12.75}                  & 4.90 \\
                             && FRLL-FaceMorpher    & 3.17          & \multicolumn{1}{c|}{30.39}  & \multicolumn{1}{c|}{\underline{0.49}}       & 0.49 \\  \hline
     \multicolumn{2}{|c|}{\multirow{5}{*}{WB-Avcivas \cite{DBLP:journals/spl/Avcibas24}}}                         
                             & FRLL-Style-GAN2      & 14.87         & \multicolumn{1}{c|}{-}  & \multicolumn{1}{c|}{-}   & - \\
                             && FRLL-WebMorph       & 19.32         & \multicolumn{1}{c|}{-}  & \multicolumn{1}{c|}{-}   & -  \\
                             && FRLL-OpenCV         & 7.91          & \multicolumn{1}{c|}{-}  & \multicolumn{1}{c|}{-}   & -\\
                             && FRLL-AMSL           & 18.23         & \multicolumn{1}{c|}{-}  & \multicolumn{1}{c|}{-}   & -\\
                             && FRLL-FaceMorpher    & 17.11         & \multicolumn{1}{c|}{-}  & \multicolumn{1}{c|}{-}   & -\\ \hline \hline
\multirow{10}{*}{MADation (ours)} & \multicolumn{1}{|c|}{\multirow{5}{*}{ViT-B}} 
                              & FRLL-Style-GAN2         & 17.21             & \multicolumn{1}{c|}{54.85}            & \multicolumn{1}{c|}{26.69}                & 13.10\\
     & \multicolumn{1}{|c|}{} & FRLL-WebMorph           & \textbf{3.42}     & \multicolumn{1}{c|}{\textbf{5.88}}    & \multicolumn{1}{c|}{\textbf{0.49}}        & \textbf{0.00}\\
     & \multicolumn{1}{|c|}{} & FRLL-OpenCV             & 2.97              & \multicolumn{1}{c|}{4.41}             & \multicolumn{1}{c|}{\underline{0.49}}     & 0.49\\
     & \multicolumn{1}{|c|}{} & FRLL-AMSL               & \textbf{3.85}     & \multicolumn{1}{c|}{\textbf{12.07}}   & \multicolumn{1}{c|}{\textbf{2.89}}        & \textbf{2.41}\\
     & \multicolumn{1}{|c|}{} & FRLL-FaceMorpher        & 1.35              & \multicolumn{1}{c|}{\underline{1.47}} & \multicolumn{1}{c|}{\textbf{0.00}}        & \textbf{0.00}\\  \cline{2-7}
 & \multicolumn{1}{|c|}{\multirow{5}{*}{ViT-L}}  
                              & FRLL-Style-GAN2         & 24.96             & \multicolumn{1}{c|}{94.17}             & \multicolumn{1}{c|}{49.03}               & 22.33\\
     & \multicolumn{1}{|c|}{} & FRLL-WebMorph           & 4.07              & \multicolumn{1}{c|}{\underline{6.86}}  & \multicolumn{1}{c|}{\underline{1.47}}    & 1.47\\
     & \multicolumn{1}{|c|}{} & FRLL-OpenCV             & \underline{0.99}  & \multicolumn{1}{c|}{\underline{0.98}}  & \multicolumn{1}{c|}{\textbf{0.00}}       & \textbf{0.00}\\
     & \multicolumn{1}{|c|}{} & FRLL-AMSL               & 7.26              & \multicolumn{1}{c|}{21.26}             & \multicolumn{1}{c|}{\underline{10.63}}   & 5.80\\
     & \multicolumn{1}{|c|}{} & FRLL-FaceMorpher        & \textbf{0.74}     & \multicolumn{1}{c|}{\textbf{0.98}}     & \multicolumn{1}{c|}{0.98}                & 0.98\\ \hline
\end{tabular}}
\vspace{-5mm}
\label{tab:frll}
\end{table}

\textbf{Comparison with the recent MAD approaches:} To further extend our study and verify if MADation shows competitive performance with recent MAD approaches, we compare our proposed framework with several MAD architectures previously proposed in the literature. The comparison considered all the reported results in the literature that complied with the SYN-MAD 2022 competition \cite{huber2022syn} by training on SMDD \cite{SMDD} and testing on the MAD22 \cite{huber2022syn} (and its derivatives \cite{damer2023mordiff})  as well as the works trained on SMDD and tested on the FRLL-Morphs \cite{DBLP:journals/corr/abs-2012-05344} evaluation benchmark. This might have missed comparisons to some published MAD techniques (that did not follow this protocol or are not publicly available) but provides a wide comparison with many of them and relies on a public and clear benchmark. In Table \ref{tab:sota}, we start by comparing the MAD techniques \cite{SMDD, damer2023mordiff, ramachandra2019detecting, DBLP:journals/corr/abs-2207-00899, huber2022syn, DBLP:conf/icb/RachalwarFDD23} with public results on MAD22 \cite{huber2022syn} and its extension MorDIFF \cite{damer2023mordiff}, including some of the submitted solutions to the SYN-MAD 2022 competition \cite{huber2022syn}. MADation presents the best and/or second-best performance in 23/21 (ViT-B/ViT-L) out of the 42 evaluated scenarios. Note that no other approach in Table \ref{tab:sota} evaluated BPCER at a fixed APCER for MorDIFF. In particular, ViT-B presents remarkable results in FaceMorph and OpenCV, while ViT-L consistently outperforms the remaining techniques in FaceMorph, MIPGAN II and WebMorph. Although the D-FW \cite{DBLP:conf/icb/RachalwarFDD23} approach presents very competitive performance levels for all available metrics, it should be kept in mind that these models use a multi-task learning framework that incorporates 3D facial information. This leads to a significant increase in computational demands and justifies the achieved performance levels. Furthermore, this information could also be included in the proposed MADation approach and would likely result in increased performance. Nonetheless, applications such as these fall out of the scope of the current work, which focuses on providing a computationally inexpensive solution to the MAD task through the usage of LoRA layers to adapt a pre-trained FM. We further extend our study to the well-known FRLL-Morphs dataset \cite{DBLP:journals/corr/abs-2012-05344} in Table \ref{tab:frll}, to present a more comprehensive comparison with previous works \cite{neto2022orthomad, caldeira2023unveiling, SMDD, DBLP:journals/spl/Avcibas24}. It can be seen that MADation scores first and/or second place in 12/9 (ViT-B/ViT-L) out of the 20 evaluated scenarios. In particular, ViT-B achieves the lowest EER for FRLL-WebMorph and FRLL-AMSL, reducing the previously best EER values by 0.59 pp. and 0.15 pp., respectively. Overall, MADation presents competitive performances with a wide set of recent MAD solutions, highlighting the importance of exploring FM's potential in biometrics tasks such as MAD.  

\vspace{-4mm}
\section{Conclusion} \label{sec:conclusion}
\vspace{-3mm}
This work presents MADation, the first approach that takes advantage of the generalization capabilities of FMs to perform MAD. To ensure that the pre-trained FM can align its feature space with the domain specificities of MAD, we adapt CLIP with LoRA layers while simultaneously training a classification layer to perform MAD. Through extensive benchmarking on several datasets and comparison with other transformer and FM-based approaches, we show that MADation can take advantage of the knowledge acquired during CLIP's pre-training with massive amounts of data while constituting a flexible approach that efficiently aligns the produced feature space with MAD specificities, resulting in increased performance. Furthermore, MADation showed competitive performance with recent MAD solutions in several evaluation benchmarks, demonstrating the potential of FM's in domain-specific tasks such as MAD provided their correct adaption to the downstream task, even when little training data is available.

\textbf{Acknowledgments:} This research work has been funded by the German Federal Ministry of Education and Research, and the Hessian Ministry of Higher Education, Research, Science and the Arts within their joint support of the National Research Center for Applied Cybersecurity ATHENE, and the Slovenian ARIS research programs Metrology and Biometric Systems (P2-0250) and Computer Vision (P2-0214).

{\small
\bibliographystyle{ieee_fullname}
\bibliography{egbib}
}

\end{document}